# Cyclic Causal Discovery from Continuous Equilibrium Data

**Joris M. Mooij**[*]
Informatics Institute
University of Amsterdam
The Netherlands

**Tom Heskes**
Institute for Computing and Information Sciences
Radboud University Nijmegen
The Netherlands

## Abstract

We propose a method for learning cyclic causal models from a combination of observational and interventional equilibrium data. Novel aspects of the proposed method are its ability to work with continuous data (without assuming linearity) and to deal with feedback loops. Within the context of biochemical reactions, we also propose a novel way of modeling interventions that modify the *activity* of compounds instead of their abundance. For computational reasons, we approximate the nonlinear causal mechanisms by (coupled) local linearizations, one for each experimental condition. We apply the method to reconstruct a cellular signaling network from the flow cytometry data measured by Sachs et al. (2005). We show that our method finds evidence in the data for feedback loops and that it gives a more accurate quantitative description of the data at comparable model complexity.

## 1 Introduction

A central question that arises in many empirical sciences is how to discover cause-effect relationships between variables from measured data. Knowledge of causal relationships is essential in order to predict how a system will react to interventions that perturb the system from its natural state (Pearl, 2000), which is very useful for many practical applications. An example from biology is the problem of predicting *in silico* how a signaling pathway in a cell will react *in vitro* when it is treated with a certain chemical compound. The ability to reliably make such causal predictions can be a powerful tool for practical applications like drug design.

A concrete example is the multivariate proteomics data set measured and analyzed by Sachs et al. (2005). Using flow cytometry, abundances of 11 biochemical compounds (phosphorylated proteins and phospholipid components) were measured in single human immune system cells under various experimental perturbations. Sachs et al. (2005) reconstructed the underlying "signaling network" by learning Bayesian networks from the data. Their reconstruction turned out to be very close to the "well-established consensus network" that had been obtained by manually combining results from many different experiments, an effort that had taken about two decades.

The consensus network contains 18 expected causal relationships. Sachs et al. (2005) found two new unexpected causal relations (and experimentally verified one of them) and obtained one reversed relationship. However, they did not recover three of the 18 expected causal relationships. Sachs et al. (2005) hypothesized that these three missing causal relationships are all involved in feedback loops. As Bayesian networks are acyclic by definition, this could explain why they were not found by their method. Additional support for this hypothesis can be found by simple inspection of the data, which already shows strong evidence for the presence of feedback loops. In later work (Itani et al., 2010), the authors proposed a heuristic method for causal discovery that takes into account the possibility of feedback. An alternative approach to dealing with cycles was proposed by Schmidt and Murphy (2009).

A common feature of the causal discovery methods that have been applied so far on this protein data set is that they all work with *discretized* data: although the raw measurements are continuous-valued, the data is preprocessed by discretization into three coarse categories (low, medium and high abundance). We argue that discretization of the data as a preprocessing step should be avoided if possible, as this throws away much

---

[*]Also affiliated to Institute for Computing and Information Sciences, Radboud University Nijmegen, The Netherlands

of the information in the data that could be useful for causal discovery. Recently, several methods have been proposed for causal discovery from continuous-valued observational data by exploiting independence of the estimated noise with the input (Shimizu et al., 2006; Hoyer et al., 2009; Zhang and Hyvärinen, 2009; Peters et al., 2011). Similar ideas have also been studied in the cyclic case (Lacerda et al., 2008; Mooij et al., 2011). More recently, cyclic methods that can deal with hidden common causes and with a combination of observational and experimental data have been proposed (Eberhardt et al., 2010; Hyttinen et al., 2012).

However, none of these methods are directly applicable to the (Sachs et al., 2005) data set, among others because they model interventions in a different way. Most interventions performed by Sachs et al. (2005) change the *activity* of a compound, not its *abundance*, and therefore the standard formalism for interventions (Pearl, 2000) is not applicable. Sachs et al. (2005) and Itani et al. (2010) propose two different ways of modeling these interventions that both exploit the fact that the data has been discretized. Eaton and Murphy (2007) consider different possible intervention types and learn the interventions from the data, instead of using the biological background knowledge in (Sachs et al., 2005). Eaton and Murphy (2007) conclude that the data can be best explained by assuming that most interventions are not as specific as originally assumed by Sachs et al. (2005), but act on multiple compounds simultaneously (also known as "fat-hand" interventions). In this work, we offer an alternative explanation, where we assume that interventions are specific (i.e., act only on a single compound), but where most interventions change the *activity* of that compound (i.e., the way in which it influences the equilibrium distributions of its direct effects). In addition, feedback loops may increase the impact of an intervention.

The goal of this work is to develop a practical method for analyzing data sets such as the protein data collected by Sachs et al. (2005). The method we propose here does not start by throwing away information (by discretizing the data as a preprocessing step), but works directly with the original continuous-valued measurements. As we expect feedback loops to play a prominent role in biological networks, we also drop the assumption of acyclicity. Another feature of our method that distinguishes it from many existing approaches is that we will not assume linearity of the causal mechanisms but allow for nonlinearities. Finally, we propose a natural and in our opinion more realistic way of modeling activity interventions.

## 2 Modeling assumptions

In this section we describe our modeling assumptions in detail. First of all, the data form a "snapshot" of a dynamical process: for each individual cell we have one multivariate measurement done at a single point in time. Therefore, we will assume that the cells have reached *equilibrium* when the measurements are performed (an assumption called "homeostasis" in biology). This is an approximation, but a necessary one in the light of the absence of time-series data.

### 2.1 Structural Causal Models

We will assume that the equilibrium data can be described by a *Structural Causal Model (SCM)* (Pearl, 2000), also known as *Structural Equation Model (SEM)* (Bollen, 1989). In particular, for $D$ observed variables $x_1, \ldots, x_D$ (corresponding in our case to the abundances of the biochemical compounds), the model consists of $D$ *structural equations*

$$x_i = f_i(\boldsymbol{x}_{\text{pa}(i)}, \epsilon_i) \qquad i = 1, \ldots, D \qquad (1)$$

where $\text{pa}(i) \subseteq \{1, \ldots, D\} \setminus \{i\}$ is the set of *parents* (direct causes) of $x_i$, $f_i$ is the *causal mechanism* determining the value of the effect $x_i$ in terms of its direct causes $\boldsymbol{x}_{\text{pa}(i)}$ and a *disturbance variable* $\epsilon_i$ representing all unobserved other causes of $x_i$. In addition, an SCM specifies a joint probability density $p(\boldsymbol{\epsilon})$ on the disturbance variables $\epsilon_1, \ldots, \epsilon_D$. Following Sachs et al. (2005), we will make the assumption of *causal sufficiency*, which means that we exclude the possibility of *confounders* (i.e., hidden common causes of two or more observed variables). In other words, we assume that the disturbance variables are jointly independent: $p(\boldsymbol{\epsilon}) = \prod_{i=1}^{n} p(\epsilon_i)$. Without loss of generality, we will additionally assume that $\mathbb{E}(\boldsymbol{\epsilon}) = \boldsymbol{0}$ and $\mathbb{V}\text{ar}(\boldsymbol{\epsilon}) = \boldsymbol{I}$.

The structure of a causally sufficient SCM $\mathcal{M}$ can be visualized with a directed graph $\mathcal{G}_\mathcal{M}$ with vertices $\{x_1, \ldots, x_D\}$ and edges $x_j \to x_i$ if and only if $f_i$ depends on $x_j$, i.e., if $j \in \text{pa}(i)$. As we do not exclude the possibility of feedback loops, the graph $\mathcal{G}_\mathcal{M}$ is not necessarily acyclic, but may contain directed cycles. We will assume that for each joint value $\boldsymbol{\epsilon}$ of the disturbance variables, there exists a unique solution $\boldsymbol{x}(\boldsymbol{\epsilon})$ of the $D$ structural equations (1). Note that this assumption is automatically satisfied in the acyclic case, but that it induces additional constraints in the cyclic case. This assumption implies that the distribution $p(\boldsymbol{\epsilon})$ induces a distribution $p(\boldsymbol{x})$ on the observed variables. This induced distribution is called the *observational distribution* of the SCM. In addition, we will assume that the mapping $\boldsymbol{\epsilon} \mapsto \boldsymbol{x}$ is invertable. In that

case, the observational density is given explicitly by:

$$p(\boldsymbol{x}) = p(\boldsymbol{\epsilon}(\boldsymbol{x})) \left|\det \frac{d\boldsymbol{\epsilon}}{d\boldsymbol{x}}\right| = \left(\prod_{i=1}^{D} p(\epsilon_i)\right) \left|\det \frac{d\boldsymbol{\epsilon}}{d\boldsymbol{x}}\right|. \quad (2)$$

## 2.2 Interventions

The SCM literature typically considers "perfect interventions", which are modeled as follows (Pearl, 2000). Under an intervention "$\mathrm{do}(x_i = \xi_i)$" that forces the variable $x_i$ to attain the value $\xi_i$, the SCM is adapted by replacing the structural equation for $x_i$ with the equation $x_i = \xi_i$, while leaving the other aspects of the SCM invariant. In particular, the distribution on the disturbance variables $p(\boldsymbol{\epsilon})$ stays the same; however, because one of the structural equations changed, the induced distribution on the observed variables $\boldsymbol{x}$ changes into the *interventional* distribution, with density $p(\boldsymbol{x} \mid \mathrm{do}(x_i = \xi_i))$. In the cyclic case, we also need to assume that under the relevant interventions, there exists a unique solution $\boldsymbol{x}(\boldsymbol{\epsilon})$ of the (modified) structural equations for each value of $\boldsymbol{\epsilon}$; otherwise, the induced (interventional) distribution will be ill-defined.

These "perfect interventions" correspond in the case of the signaling network data with interventions that change the *abundance* of a compound. However, many of the interventions actually performed by (Sachs et al., 2005) do not directly change the abundance, but rather its *activity*, i.e., the extent to which it influences abundances of other compounds. In their original paper, (Sachs et al., 2005) model these "activity interventions" in the following way: if the activity of compound $i$ is *inhibited*, the actual measurements of $x_i$ are replaced with the value "low", whereas if compound $i$ is *activated*, the actual measurements of $x_i$ are replaced with the value "high". Not only does this approach throw away data, it also depends on the discretization of the data. In later work, (Itani et al., 2010) model these interventions in a different way: they split the variable $x_i$ into two parts, $x_i$ and $x_i^{\mathrm{int}}$, where $x_i^{\mathrm{int}}$ is assigned the value corresponding to the intervention (either "low" in case of an inhibitor or "high" in case of an activator), and $x_i$ represents the abundance of compound $i$ measured in the interventional experiment. In the modified graph corresponding to the intervention, the outgoing arrows from $x_i$ now become outgoing arrows of $x_i^{\mathrm{int}}$ instead, and all incoming arrows go into $x_i$. This approach no longer throws away data, but it still requires a coarse discretization of the data.

Instead, we propose to model these activity interventions as follows: if an intervention changes the *activity* of compound $i$, we adapt the SCM by allowing the *children*[1] $j$ of compound $i$ to change their causal mechanism $f_j(\boldsymbol{x}_{\mathrm{pa}(j)}, \epsilon_j)$ into a different function $\tilde{f}_j(\boldsymbol{x}_{\mathrm{pa}(j)}, \epsilon_j)$, whereas the other aspects of the SCM (including its structure) remain invariant. In our context, this new causal mechanism $\tilde{f}_j$ is unknown and we learn it from the data. In particular, we do not use the background knowledge provided by Sachs et al. (2005) that specifies whether an activity intervention is an inhibitor or an activator.

## 2.3 Approximating causal mechanisms

So far, we have not assumed linearity, and in theory we could proceed by modeling the causal mechanisms as nonparametric nonlinear functions, e.g., as Gaussian Processes (Rasmussen and Williams, 2006). For computational reasons, however, we linearize the causal mechanisms in the SCM locally around their average input $(\langle \boldsymbol{X}_{\mathrm{pa}(i)}\rangle, 0)$:[2]

$$f_j(\boldsymbol{x}_{\mathrm{pa}(j)}, \epsilon_j) \approx \sum_{i=1}^{D} B_{ij} x_i + \mu_j + \alpha_j \epsilon_j$$

where we introduced the matrix $\boldsymbol{B} \in \mathbb{R}^{D \times D}$ and vectors $\boldsymbol{\mu}, \boldsymbol{\alpha} \in \mathbb{R}^{D \times 1}$, defined by:

$$B_{ij} := \left.\frac{\partial f_j}{\partial x_i}\right|_{(\langle \boldsymbol{X}_{\mathrm{pa}(j)}\rangle, 0)}, \quad \alpha_j := \left.\frac{\partial f_j}{\partial \epsilon_j}\right|_{(\langle \boldsymbol{X}_{\mathrm{pa}(j)}\rangle, 0)},$$

$$\mu_j := f_j(\langle \boldsymbol{X}_{\mathrm{pa}(j)}\rangle, 0) - \sum_{i \in \mathrm{pa}(j)} \left.\frac{\partial f_j}{\partial x_i}\right|_{(\langle \boldsymbol{X}_{\mathrm{pa}(j)}\rangle, 0)} \langle X_i \rangle.$$

Note that the structure of the matrix $\boldsymbol{B}$ reflects the graph structure $\mathcal{G}_{\mathcal{M}}$ of the model: it has zeroes on the diagonal, and $B_{ij}$ is the (linearized) direct effect of $x_i$ on $x_j$, which can only be nonzero if $i \in \mathrm{pa}(j)$.

This means that for a single experimental condition, we assume the following linearized structural equations:

$$x_j = \boldsymbol{x}^T \boldsymbol{B}_{\cdot j} + \mu_j + \alpha_j \epsilon_j.$$

For an i.i.d. sample of $N$ data points arranged in the matrix $\boldsymbol{X} \in \mathbb{R}^{N \times D}$ and latent disturbance variables $\boldsymbol{E} \in \mathbb{R}^{N \times D}$, these can be written in matrix notation:

$$\boldsymbol{X}(\boldsymbol{I} - \boldsymbol{B}) = \boldsymbol{1}_N \boldsymbol{\mu}^T + \boldsymbol{E}\boldsymbol{\alpha}^T. \quad (3)$$

Under some upstream intervention, the average input $(\langle \boldsymbol{X}_{\mathrm{pa}(i)}\rangle, 0)$ of the causal mechanism $f_i$ for compound $i$ may change. If the change is large with respect to the curvature of $f_i$, we may need to relinearize $f_i$ around the new average input under the intervention (even though the nonlinear function $f_i$ itself may have remained unchanged, see also Figure 2(a)). This will be discussed in detail in section 2.5.

---
[1] The children of $i$ are all $j$ such that $i \in \mathrm{pa}(j)$.

[2] We denote the empirical mean of a variable $x$ by $\langle X \rangle$.

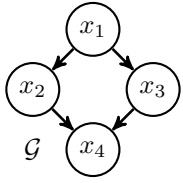

| $c$ | Intervention type |
|---|---|
| 1 | none (observational) |
| 2 | activity of $x_1$ |
| 3 | activity of $x_2$ |
| 4 | abundance of $x_3$ |
| 5 | abundance of $x_1$ |

$$m_{ic}(\mathcal{G}) = \begin{pmatrix} 1 & 1 & 1 & 1 & 2 \\ 1 & 2 & 1 & 1 & 1 \\ 1 & 2 & 1 & 3 & 1 \\ 1 & 1 & 2 & 1 & 1 \end{pmatrix} \quad M_i(\mathcal{G}) = \begin{pmatrix} 1 \\ 2 \\ 3 \\ 2 \end{pmatrix}$$

Figure 1: Example of a graph $\mathcal{G}$, experimental metadata, and the corresponding mechanism labels $m_{ic}(\mathcal{G})$. As an example, $m_{32}(\mathcal{G}) = 2$ because the second experimental condition, an activity intervention on $x_1$, changes the causal mechanism of $x_3$. In the third condition, the causal mechanism of $x_3$ is identical to that in the first condition, so $m_{33}(\mathcal{G}) = m_{31}(\mathcal{G}) = 1$.

## 2.4 Likelihood

Assuming that all disturbance variables $\epsilon_i$ have the same probability density $p(\epsilon_i) = p_0$, the likelihood of i.i.d. data $\boldsymbol{X}$ for a single experimental condition follows directly from expressions (2) and (3):

$$p(\boldsymbol{X} \mid \boldsymbol{B}, \boldsymbol{\mu}, \boldsymbol{\alpha}) = \prod_{n=1}^{N} \left[ |\det(\boldsymbol{I} - \boldsymbol{B})| \cdot \prod_{i=1}^{D} \frac{1}{\alpha_i} p_0 \left( \frac{(\boldsymbol{X}(\boldsymbol{I} - \boldsymbol{B}))_{ni} - \mu_i}{\alpha_i} \right) \right]. \quad (4)$$

We will consider two choices for the noise density, Gaussian noise $p_0(e) = \frac{1}{\sqrt{2\pi}} \exp(-\frac{1}{2}e^2)$ and super-Gaussian noise that is often used in the Independent Component Analysis (ICA) literature, $p_0(e) = 1/(\pi \cosh(e))$. Note that in the acyclic case, $\det(\boldsymbol{I} - \boldsymbol{B}) = 1$, and therefore the likelihood factorizes over variables. This simplification does not occur in the cyclic case, as the likelihoods of different variables may become coupled through the determinant.

Combining data from different experimental conditions $c = 1, \ldots, K$ is straightforward:

$$p\big((\boldsymbol{X})_{c=1}^{K} \mid (\boldsymbol{B}^{(c)}, \boldsymbol{\mu}^{(c)}, \boldsymbol{\alpha}^{(c)})_{c=1}^{K}\big)$$
$$= \prod_{c=1}^{K} p(\boldsymbol{X}^{(c)} \mid \boldsymbol{B}^{(c)}, \boldsymbol{\mu}^{(c)}, \boldsymbol{\alpha}^{(c)}),$$

where the superscript "$(c)$" labels data and parameters corresponding to the $c$'th experimental condition.

## 2.5 Parameter priors

We denote all parameters of the linearized SCM corresponding to experimental condition $c$ by $\boldsymbol{\Theta}^{(c)} :=$

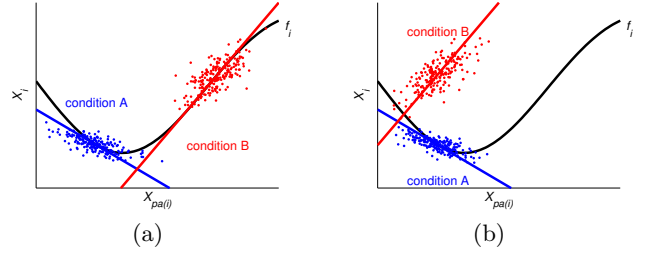

Figure 2: (a) Even though some causal mechanism $f_i(\boldsymbol{x}_{\text{pa}(i)}, \epsilon_i)$ stays invariant, its linearization around the new equilibrium may have changed. (b) In this case, even a nonlinear causal mechanism cannot fit the data well. Another structure, assigning different causal mechanisms to conditions A and B, may better fit the data.

$(\boldsymbol{B}^{(c)}, \boldsymbol{\mu}^{(c)}, \log \boldsymbol{\alpha}^{(c)})$, and the subset of parameters corresponding only to the causal mechanism $f_i^{(c)}$ of the $i$'th compound as $\boldsymbol{\Theta}_i^{(c)} := (B_{\cdot i}^{(c)}, \mu_i^{(c)}, \log \alpha_i^{(c)})$. Using the Bayesian approach to multi-task learning, we couple these $K$ learning problems by imposing a prior $p(\boldsymbol{\Theta}^{(1)}, \ldots, \boldsymbol{\Theta}^{(K)} \mid \mathcal{G})$ on the parameters that embodies our assumption that these parameters should be related in specific ways, conditional on the hypothetical causal structure $\mathcal{G}$ of the SCM. The hypothetical structure $\mathcal{G}$ always constrains the structure of $\boldsymbol{B}$ in the sense that if $i \notin \text{pa}_{\mathcal{G}}(j)$, $B_{ij}^{(c)} = 0$ for all $c = 1, \ldots, K$. We consider various choices to couple the nonzero parameters of the $\boldsymbol{B}^{(c)}$, and the location and scale parameters $\boldsymbol{\mu}^{(c)}$ and $\boldsymbol{\alpha}^{(c)}$, across experimental conditions.

For each compound $i$, the prior introduces couplings between $\boldsymbol{\Theta}_i^{(c)}$ for all conditions $c$ which have the same causal mechanism (i.e., if $f_i^{(c')} = f_i^{(c)}$). An intervention may change $f_i$ into another function $\tilde{f}_i$; whether or not such a mechanism change happens, depends on the experimental condition $c$ and on the hypothetical causal structure $\mathcal{G}$ (remember that an abundance intervention on compound $i$ changes its causal mechanism $f_i$ and an activity intervention on compound $i$ changes the causal mechanisms of its children $\{f_j\}_{j \in \text{ch}_{\mathcal{G}}(i)}$). Let the causal mechanism for compound $i$ in condition $c$ be given by $f_i^{(c)} = \phi_{i, m_{ic}(\mathcal{G})}$, where the label $m_{ic}(\mathcal{G}) \in \{1, 2, \ldots, M_i(\mathcal{G})\}$ depends on the causal structure $\mathcal{G}$ (see also Figure 1). Here, $M_i(\mathcal{G})$ is the total number of different causal mechanisms for compound $i$ needed to account for all experimental conditions. We take a prior that couples parameters corresponding to the same causal mechanisms:

$$p(\boldsymbol{\Theta} \mid \mathcal{G}) = \prod_{i=1}^{D} \prod_{m=1}^{M_i(\mathcal{G})} p\big((\boldsymbol{\Theta}_i^{(c)})_{m_{ic}(\mathcal{G}) = m} \mid \mathcal{G}\big).$$

Note that this prior couples parameters $\boldsymbol{\Theta}_i^{(c)}$ with $\boldsymbol{\Theta}_i^{(c')}$ only if $m_{ic}(\mathcal{G}) = m_{ic'}(\mathcal{G})$.

We will consider two different choices for the factors $p\big((\boldsymbol{\Theta}_i^{(c)})_{m_{ic}(\mathcal{G})=m} \,|\, \mathcal{G}\big)$, corresponding to different degrees of approximation of the fact that the parameters $\{\boldsymbol{\Theta}_i^{(c)}\}_{c=1}^K$ correspond with linearizations of the latent nonlinear causal mechanisms $\{\phi_{i,m}\}_{m=1}^{M_i(\mathcal{G})}$.

### 2.5.1 Linear mechanisms prior

This prior assumes that no relinearizations of the descendants of an intervention node are required. In other words, if one or more causal mechanisms change as a result of some intervention, the input distributions of the descendant variables are assumed to change not too much, such that their linearization remains approximately the same. We can then use hard equality constraints:

$$p\big((\boldsymbol{\Theta}_i^{(c)})_{m_{ic}(\mathcal{G})=m} \,|\, \mathcal{G}\big)$$
$$= \int p(\boldsymbol{\Theta}_i^m) \prod_{\substack{c=1 \\ m_{ic}(\mathcal{G})=m}}^{K} \delta(\boldsymbol{\Theta}_i^{(c)} - \boldsymbol{\Theta}_i^m) \, d\boldsymbol{\Theta}_i^m$$

with

$$p\big(\boldsymbol{\Theta}_i^m = (\boldsymbol{b}, \mu, a)\big)$$
$$= \mathcal{N}(\boldsymbol{b} \,|\, \boldsymbol{0}_D, \lambda^2 \mathrm{diag}(\mathcal{G}_{i,\cdot})) \, \mathcal{N}(\mu \,|\, 0, \tau) \, \mathcal{N}(a \,|\, 0, \tau)$$

where $a = \log \alpha$ and where we let $\tau \to \infty$, which yields a flat prior over the location $\mu_i$ and Jeffrey's prior over the scale $\alpha_i$. We have a single hyperparameter $\lambda$ for penalizing the nonzero components of $\boldsymbol{b} = \boldsymbol{B}_{\cdot i}^{(c)}$.

### 2.5.2 Nonlinear mechanisms prior

The previous prior does not deal well with the situation in Figure 2(a). Here, condition A could be the baseline (observational condition), and condition B could be an intervention that changes something upstream of $\boldsymbol{x}_i$, but keeps the mechanism $f_i$ unchanged. Because the upstream intervention may lead to a change in input distribution of the parents $\boldsymbol{x}_{\mathrm{pa}(i)}$, relinearization of $f_i$ around a new average input is desirable in general. Therefore, we introduce a prior that allows for downstream relinearizations. We have tried a prior that allows *all* descendants of an intervention target in condition $c \neq 1$ to pick parameters $\boldsymbol{\Theta}_i^{(c)}$ *independent* of the baseline parameters $\boldsymbol{\Theta}_i^{(1)}$ in the observational setting $c = 1$. That prior does yield better results in the acyclic case than the prior in Section 2.5.1, but in the cyclic case it leads to "cheating" in the sense that the prior strongly encourages to introduce one big directed cycle that connects all the variables. Then, each variable is a descendant of each other variable, and can pick new (independent) parameters in *each* experimental condition, effectively completely decoupling all experimental conditions.

The solution we propose here is a compromise that replaces the hard equality constraints of the prior in section 2.5.1 by soft constraints. The idea is to model each causal mechanism $f_i^{m_{ic}(\mathcal{G})}(\boldsymbol{x}_{\mathrm{pa}(i)}, \epsilon_i)$ as a Gaussian Process (GP) and interpret the parameters $(B_{\cdot i}^{(c)}, \mu_i^{(c)}, \alpha_i^{(c)})$ as pseudo-data for the GP (Solak et al., 2003). They note that for Gaussian Process regression, one is not necessarily restricted to using pairs of input and output, but one can combine this data with data regarding the derivative of the output with respect to some input dimension, at a given input location. In our case, the "data" are actually the linearized parameters $(B_{\cdot i}^{(c)}, \mu_i^{(c)}, \alpha_i^{(c)})$, which are coupled to the real data via the likelihood (4). We use an isotropic squared exponential covariance function:

$$k\big((\boldsymbol{x}_{\mathrm{pa}(i)}, \epsilon_i), (\tilde{\boldsymbol{x}}_{\mathrm{pa}(i)}, \tilde{\epsilon}_i)\big)$$
$$= \sigma_{\mathrm{out}}^2 \exp\left(-\frac{(\boldsymbol{x}_{\mathrm{pa}(i)} - \tilde{\boldsymbol{x}}_{\mathrm{pa}(i)})^2}{2\sigma_{\mathrm{in}}^2}\right) \exp\left(-\frac{(\epsilon_i - \tilde{\epsilon}_i)^2}{2\sigma_{\mathrm{in}}^2}\right)$$

and add a small "jitter" term for numerical stability purposes (i.e., we add $\sigma_{\mathrm{jitter}}^2 \boldsymbol{I}$ to the kernel matrix $\boldsymbol{K}$). Similar to the prior in 2.5.1, this GP prior couples different $c$ for the same $i$. As the determinant factor in the likelihood couples different $i$ for the same $c$, we cannot simply use the trick of Solak et al. (2003) (who use the posterior distribution of the biases and slopes of Bayesian linear regressions as pseudo-data), but have to apply a more global approximation scheme (see Section 2.6).

This prior deals well with the situation in Figure 2(a), as the pseudo-data corresponding to the two local linear models would have a high probability under this GP prior. On the other hand, the GP prior strongly penalizes situations such as in Figure 2(b), in line with our intuition that the same causal mechanism $f_i$ cannot be a good model for the data of both condition A and B in that case.

### 2.6 Structure priors and scoring structures

We use an approximate Bayesian approach to calculate the posterior probability of a putative causal graph $\mathcal{G}$, given the data and prior assumptions. In principle, exact Bayesian scoring would yield automatic regularization (if our assumption that there is no confounding holds true). However, as the posterior distribution is intractable, we have to approximate it. Given a hypothetical causal structure $\mathcal{G}$, we numerically optimize the posterior with respect to the parameter and employ the Laplace approximation (Laplace, 1774) to get an approximation of the evidence (marginal likelihood) for that structure.

The number of possible causal graphs $\mathcal{G}$ grows very quickly as a function of the number of variables: for the Sachs et al. (2005) data, which has $D = 11$ variables, there are about $3.1 \times 10^{22}$ different directed acyclic graphs (DAGs) and $2^{D^2-D} \approx 1.2 \times 10^{33}$ directed graphs. Even though calculating the evidence for a single structure is doable, exhaustive enumeration or scoring is clearly hopeless. Therefore, we use greedy optimization methods (local search) in the hope to find the important modes of the posterior over causal structures. We use simple priors over structures: a flat prior over directed graphs, a flat prior over acyclic graphs, and flat priors over all graphs (either acyclic or all directed graphs) that have at most $n$ edges.

If exact Bayesian inference were feasible, we could either select the best scoring structure, or average over structures according to their evidence, in order to obtain predictions. However, as we are using approximate inference, we will also use *stability selection* (Meinshausen and Bühlmann, 2010) to assess the stability of posterior edge probabilities.

## 3 Application on real-world data

In this section, we describe the results of our proposed method on the flow cytometry data set.

### 3.1 Properties of the data

The data published by Sachs et al. (2005) is a good test case for causal discovery methods for several reasons. First, the high quality of the data:[3] each sample is a multivariate measurement in a *single cell* (usually, only population averages are measured), the number of data points is large (about $10^4$ in total), and the measurement noise seems to be relatively low. Furthermore, knowledge about the "ground truth" is available, which helps verification of results. Finally, good results have already been demonstrated with acyclic causal discovery methods, but the data is interesting for our purposes as it shows evidence of feedback relationships.

Figure 3(a) shows a subset of the data as a heat map. Table 1 describes the biological background knowledge about the different experimental conditions: which reagent has been added, and what is the known effect of this reagent? We used a subset of 8 of the available 14 experimental conditions. Figure 3(b) shows whether the interventional distributions are significantly different from the observational distribution, for each variable and each experimental condition. Figure 4 shows two scatter plots of the data in two different experimental conditions. Note the almost perfect linear relationship between log-abundance of Raf and Mek in condition 5, which implies that the measurement noise (i.e., the noise added by the measurement device) must be relatively small. This also shows a strong dependence between Raf and Mek, which is expected from the consensus network (where Raf is a direct cause of Mek). On the other hand, note the absence of dependence between Mek and Erk. Assuming the consensus (Mek causes Erk) to be true, this is an example of a *faithfulness* violation. The data actually shows more such faithfulness violations, which makes causal discovery challenging (but not necessarily impossible, since we do have interventional data). Furthermore, note that the intervention on Mek (condition 5) changes the Raf concentration. So, assuming that the consensus that Raf causes Mek is true, this is an example of feedback. Another example of feedback is that changing the activity of Mek results in a change of abundance of Mek itself.[4] Finally, the figure shows one more aspect of the data: it is censored by the detection limit of the measurement device (i.e., all abundances lower than some threshold $\theta = 1$ are assigned the value $\theta$).

Table 1: Experimental metadata: conditions we used for inferring the causal structure. The information about the type of intervention is used as background knowledge for causal discovery.

| $c$ | Reagent | Intervention |
|---|---|---|
| 1 | - | none (observational) |
| 2 | Akt-inhibitor | inhibits AKT activity |
| 3 | G0076 | inhibits PKC activity |
| 4 | Psitectorigenin | inhibits PIP2 abundance |
| 5 | U0126 | inhibits MEK activity |
| 6 | LY294002 | changes PIP2/PIP3 mechanisms |
| 7 | PMA | activates PKC activity |
| 8 | $\beta$2CAMP | activates PKA activity |

### 3.2 Results

The consensus network and the reconstruction by Sachs et al. (2005) are illustrated in Figure 5.

We experimented with several different combinations of structure and parameter priors. We used hyperparameter $\lambda = 10$ for the linear mechanisms prior (Section 2.5.1), and $\sigma_{\text{in}} = \sigma_{\text{out}} = 10$ for the nonlinear

---

[3] However, we did discover an error in the published data: the first 848 measurements of RAF and MEK in the third experimental condition (AKT-inhibitor) are identical to those in the seventh condition (LY294002). We informed the authors about this and decided to ignore this issue here.

[4] An alternative explanation of these observations could be non-specificity of the intervention reagents (Eaton and Murphy, 2007).

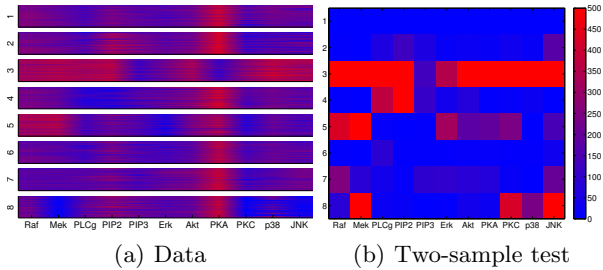

(a) Data　　　　(b) Two-sample test

Figure 3: (a) Subset of the data from Sachs et al. (2005). Color corresponds with log-abundance (red is high, blue is low); columns correspond with compounds (phosphorylated proteins and phospholipids); numbered subsets correspond with different experimental conditions (see also Table 1); lines within a row correspond with data samples (i.e., individual cells). (b) Negative log $p$-value of the Kolmogorov-Smirnov two-sample test, comparing the data of condition $c$ (on the vertical axis) with the observational data (condition $c = 1$). Color indicates how significantly different the two distributions are (red meaning that the difference is extremely significant).

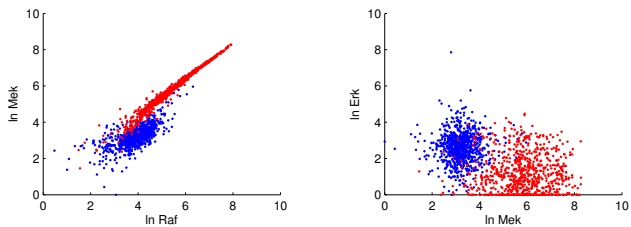

Figure 4: Scatter plot of log abundances of Mek vs. Raf (left) and Erk vs. Mek (right). Blue: condition 1 (no intervention); Red: condition 5 (MEK inhibitor).

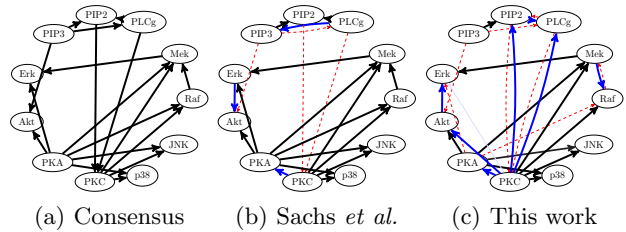

(a) Consensus　　(b) Sachs et al.　　(c) This work

Figure 5: (a) Consensus network, according to Sachs et al. (2005); (b) Reconstruction of the signaling network by Sachs et al. (2005), in comparison with the consensus network; (c) Our best *acyclic* reconstruction with at most 17 edges. Black edges: expected. Blue edges: unexpected, novel findings. Red dashed edges: missing.

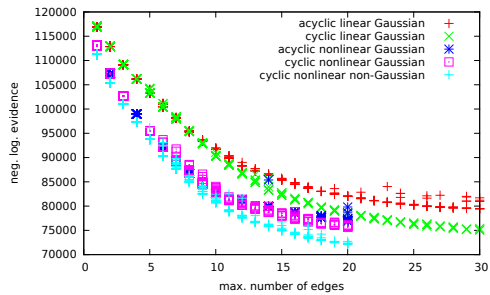

Figure 6: Negative log-evidence as a function of the maximum number of edges. Each point is a local optimum with respect to structures.

mechanisms prior (Section 2.5.2), with $\sigma_{\text{jitter}} = 0.01$. Using smaller values of the jitter did not yield significantly different results, but increased computation time considerably. Figure 6 shows how the log-evidence depends on $n$, the maximum number of edges. Each point in the plot is the result of a new greedy optimization from a different random starting point. Especially for higher numbers of edges, local maxima over structures are present, but we often seem to find the global maximum with only a few restarts of the local search procedure. Our stability selection results with a constraint on the maximum number of edges are shown in Figure 5(c) (with acyclicity constraint) and Figure 7 (cycles allowed).

In the strongly regularized acyclic case (Figure 5(c)) the precise form of the multitask prior is not very relevant: almost identical results are obtained with the (non)linear prior and/or (non-)Gaussian noise (not shown). The selected edges are very robust. Notice that our reconstruction shows less similarity with the consensus network than the reconstruction of Sachs et al. (2005) (cf. Figure 5(b)). However, when looking more closely at the unexpected edges in our acyclic reconstruction, one sees that they actually explain the data quite well. For example, our finding that Mek causes Raf (instead of vice versa) is consistent with the strong change in Raf abundance due to the Mek inhibitor (condition 5, see also Figure 4 and Figure 3(b)).[5] Similarly, the other unexpected edges in our reconstruction can all be understood qualitatively by combining the information in Figure 3(b) with that in Table 1.

When allowing for cycles, the dependence on the prior is more noticeable (see Figure 7). Nevertheless, there is reasonable agreement between the results for different priors. We see evidence for three two-cycles: Mek ⇄ PKC, Akt ⇄ Erk and Mek ⇄ PKA. When regularizing less strongly by increasing $n$ (the number of edges), re-sults become more prior dependent. There also seems to be some evidence for a two-cycle PIP2 ⇌ PLCg.

---

[5]Given this strong effect, it is surprising that Sachs et al. (2005) do find the opposite arrow. Presumably this is due to the fact that they are using information about the sign of the activity intervention (i.e., whether it is an activator or an inhibitor).

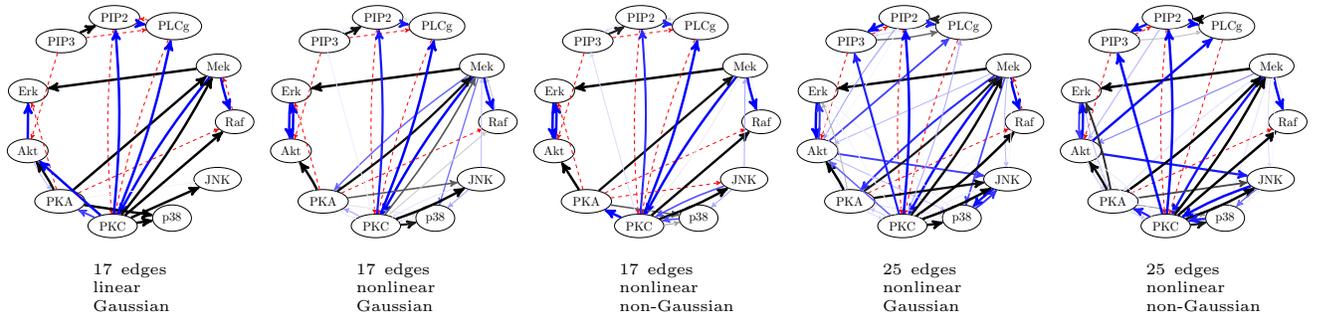

Figure 7: Stability selection results with a constraint on the number of edges, for various priors. Edge thickness and intensity reflect the probability of selecting that edge in the stability selection procedure.

Table 2: Negative log-evidences of our estimated structures (with max. 17 edges) for various structure and parameter priors in comparison with the negative log-evidences of the consensus structure and optimal structure found by Sachs et al. (2005) with the same parameter prior. All values are in units of $10^3$.

| Structure & Parameter Prior | Consensus | Sachs | This work |
|---|---|---|---|
| Acyclic, linear, Gaussian | 96.5 | 92.0 | **83.7** |
| Cyclic, linear, Gaussian | 96.6 | 92.1 | **80.4** |
| Acyclic, nonlinear, Gaussian | 87.8 | 81.8 | **77.7** |
| Cyclic, nonlinear, Gaussian | 87.8 | 81.8 | **76.6** |
| Cyclic, nonlinear, non-Gaussian | 85.4 | 79.2 | **72.9** |

In the acyclic case, parameter estimates conditional on graph structure are very robust. In the cyclic case, this no longer holds, and parameters can often not be estimated reliably from the data (as can be concluded from their posterior variance according to the Laplace approximation, but also from the lack of robustness of their estimates and the occurence of local maxima of the posterior parameter distribution). Empirically, we observed that the *structure* of the estimated graph is much more robust, though.

In Table 2 we compare the scores of some of our structures with the score of the consensus structure and that of the reconstruction by Sachs et al. (2005). Unsurprisingly, our scores are always at least as good (because they result from an optimization of scores over structures, whereas the other structures are fixed), but in all cases, the improvement is considerable.

## 4 Discussion

Performing a proper causal analysis of the (Sachs et al., 2005) data is a challenging task for various reasons. First of all, time-series data are absent, so we can only work under the equilibrium assumption. Both confounders and feedback loops are expected to be present. Most of the interventions cannot be appropriately modeled with the standard formalism, the "do-operator" (Pearl, 2000), but need to be modeled in another way. Furthermore, assumptions about the specificity of interventions may be unrealistic. Finally, several strong faithfulness violations seem to be present. This work addresses several of these issues.

Our analysis confirms the hypothesis that several feedback loops are present in the underlying system. We showed that our method gives a more accurate quantitative description of the data at comparable model complexity compared to existing methods. An interesting question from the causal point of view is whether or not our method also gives more accurate predictions for the effects of unseen interventions. We hope to address this question in the future. However, it is likely that it can only be answered definately by carrying out additional validation experiments.

We observed empirically that in the cyclic case, the parameters are often not identifiable, even though the structure is. This observation has important implications for the ability to make predictions for unseen interventions: even though reliable qualitative predictions seem possible (e.g., "an intervention on $x_i$ has (no) effect on $x_j$"), quantitative predictions depend strongly on the parameter estimates. As the parameters cannot be estimated reliably from this data, the quantitative predictions will be unreliable as well. This does not mean that making such quantitative predictions is hopeless in principle, though. Indeed, the alternative conclusion could simply be that more experimental data is needed in order to do so reliably.

In future work, we plan to compare our local linearization approach with other approximations, e.g., FITC (Snelson and Ghahramani, 2006). Also, a way to take into account the information about the sign of the activity intervention may further improve the results. Finally, we hope to find collaborators for experimental validation of our findings.

## Acknowledgements

We thank Bram Thijssen, Tjeerd Dijkstra and Tom Claassen for stimulating discussions. We also thank Karen Sachs for kindly answering our questions about the data. JM was supported by NWO, the Netherlands Organization for Scientific Research (VENI grant 639.031.036).